\documentclass[10pt,twocolumn,letterpaper]{article}

\usepackage[pagenumbers]{iccv} 

\usepackage{graphicx}
\usepackage{amsmath}
\usepackage{amssymb}
\usepackage{booktabs}
\usepackage{lipsum}

\usepackage{setspace}
\definecolor{backcolour}{rgb}{0.97,0.97,0.97}
\usepackage{listings}
\lstdefinestyle{markdownstyle}{
    basicstyle=\ttfamily\tiny,
    backgroundcolor=\color{backcolour},   
    xleftmargin=0.05\textwidth,
    xrightmargin=0.05\textwidth,
    breakindent=0\dimen0,
    columns=flexible,
    showspaces=false,
    showstringspaces=false,
    breaklines=true,
    breakatwhitespace=true,
    breakautoindent=true,
}

\definecolor{iccvblue}{rgb}{0.21,0.49,0.74}
\usepackage[pagebackref,breaklinks,colorlinks,allcolors=iccvblue]{hyperref}

\def\paperID{11591} 
\def\confName{ICCV}
\def\confYear{2025}

\title{LuciBot: Automated Robot Policy Learning from Generated Videos} 

\author{
Xiaowen Qiu\footnotemark[1] \\
  Umass Amherst\\
  \texttt{xiaowenqiu@umass.edu} \\
  \and
  Yian Wang\thanks{Equal Contribution} \\
  Umass Amherst\\
  \texttt{yianwang@umass.edu} \\
  \and
  Jiting Cai\footnotemark[1] \\
  Shanghai Jiao Tong University\\
  \texttt{caijiting@sjtu.edu.cn} \\
  \and
  Zhehuan Chen \\
  Umass Amherst\\
  \texttt{zhehuanchen@umass.edu} \\
  \and
  Chunru Lin \\
  Umass Amherst\\
  \texttt{chunrulin@mit.edu} \\
  \and
  Tsun-Hsuan Wang \\
  MIT\\
  \texttt{tsunw@mit.edu} \\
  \and
  Chuang Gan \\
  Umass Amherst\\
  \texttt{chuanggan@umass.edu} \\
}

\begin{document}
\maketitle
\begin{abstract}


Automatically generating training supervision for embodied tasks is crucial, as manual designing is tedious and not scalable. While prior works use large language models (LLMs) or vision-language models (VLMs) to generate rewards, these approaches are largely limited to simple tasks with well-defined rewards, such as pick-and-place. This limitation arises because LLMs struggle to interpret complex scenes compressed into text or code due to their restricted input modality, while VLM-based rewards, though better at visual perception, remain limited by their less expressive output modality. To address these challenges, we leverage the imagination capability of general-purpose video generation models. Given an initial simulation frame and a textual task description, the video generation model produces a video demonstrating task completion with correct semantics. We then extract rich supervisory signals from the generated video, including 6D object pose sequences, 2D segmentations, and estimated depth, to facilitate task learning in simulation. Our approach significantly improves supervision quality for complex embodied tasks, enabling large-scale training in simulators.\footnote{Project page: \url{https://wangyian-me.github.io/LuciBot/}}

\end{abstract}    
\section{Introduction}
\label{sec:intro}

Large-scale synthetic data collection has recently emerged as a crucial direction in embodied AI research~\cite{ha2023scaling, wang2023gen, wang2023robogen, dalal2023imitating, yang2024holodeck, wang2025architect, qiu2025articulate}. As demonstrated in previous works~\cite{wang2023robogen, wang2023gen, kwon2024language, ma2023eureka, rocamonde2023vision, mahmoudieh2022zero, wang2024rl, venuto2024code, yang2024robot, sontakke2024roboclip, ma2024vision}, automatically generating training supervision (reward or cost functions) based on a task description is a key component in pipelines designed to autonomously collect demonstration action trajectories for solving tasks. This capability enables skill acquisition without human intervention through training with the generated supervision using methods, such as reinforcement learning, trajectory optimization, evolution strategies, etc. Recent efforts have explored multiple approaches to tackle this problem: directly utilizing large language models to generate reward code given predefined APIs~\cite{chu2023accelerating, triantafyllidis2024intrinsic, kwon2024language, ma2023eureka, xietext2reward, yu2023language, wang2023robogen, venuto2024code}, computing text-image scores to derive rewards based on rendered states~\cite{rocamonde2023vision, mahmoudieh2022zero, sontakke2024roboclip}, and employing large vision-language models (e.g., GPT-4o) to rank rendered states~\cite{wang2024rl}. 

\begin{figure}[t]
    \centering
    \includegraphics[width=\linewidth]{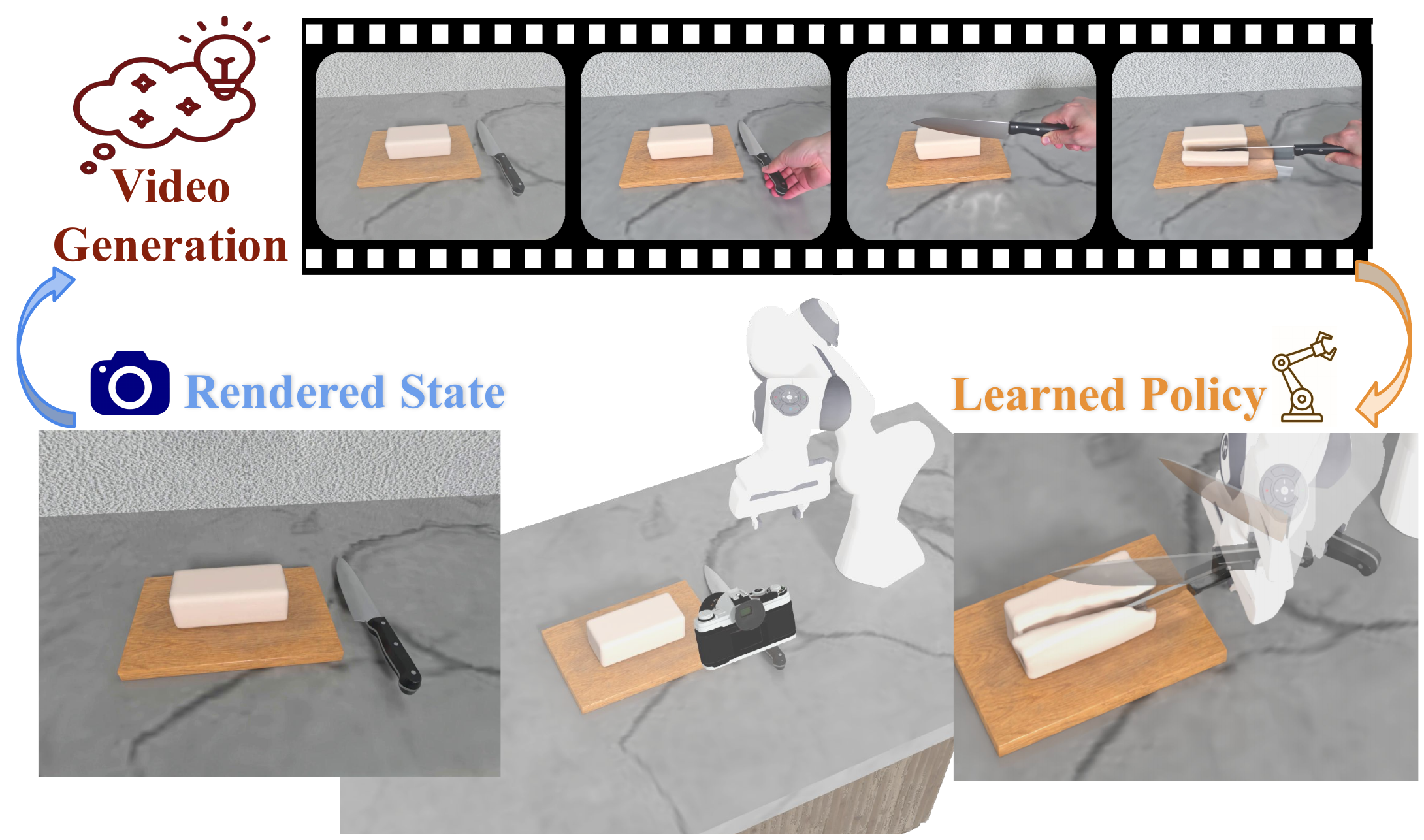}
    \vspace{-4mm}
    \caption{We present \textit{LuciBot}, a framework that utilizes a general-purpose video generation model to obtain rich supervision signals. Leveraging the semantic guidance of video generation model, \textit{LuciBot} extract rich supervision to facilitate complicated embodied AI tasks.}
    \vspace{-5mm}
    \label{fig:method}
\end{figure}

However, existing methods still face significant limitations. Language models and predefined APIs primarily operate on the underlying object states without visual input~\cite{chu2023accelerating, triantafyllidis2024intrinsic, kwon2024language, ma2023eureka, xietext2reward, yu2023language, wang2023robogen}, making them inadequate for complex cases. For example, in the \textit{Pour Liquid} task, a text-to-reward approach might maximize the fraction of liquid within the cup’s bounding box but fail to penalize spillage or improper actions, such as inserting the bottle into the cup. Additionally, text-based approaches struggle to provide rich supervision, as accurately describing liquid flow using natural language is challenging.

Methods that generate 3D goal states (e.g., RoboGen~\cite{wang2023robogen}) perform well for tasks like dough formation but struggle to represent more complex goals, such as defining the 3D goal for scooped sand or accurately modeling a cut dough.  
Some approaches~\cite{rocamonde2023vision, mahmoudieh2022zero, sontakke2024roboclip} leverage vision-language models (VLMs) like CLIP to directly measure rendered states and provide scalar reward signals. However, these methods are often imprecise and struggle to capture fine-grained differences between similar states.  
Other works, such as RL-VLM-F~\cite{wang2024rl}, employ general-purpose VLMs to rank rendered states, but these approaches require extensive VLM queries during training, making them computationally expensive and slow. Moreover, VLM-based rewards lack the accuracy needed for complex manipulation tasks, as they output textual descriptions or learned embeddings, which are often insufficient to represent high-dimensional object states, particularly for soft objects. Finally, prior attempts to use video generation models as reward functions\cite{escontrela2023video, huang2024diffusion, luo2024grounding} or policies\cite{liang2024dreamitate, ajay2023compositional, du2023learning, ko2023learning} require fine-tuning on domain-specific data, severely limiting generalizability.

As a result, solving intricate tasks where the reward is ambiguous or difficult to directly represent using APIs remains an open challenge. To address this problem, we propose leveraging a general-purpose, text-image-conditioned video generation model~\cite{kling} to imagine both the goal state and the action process required to complete the task, conditioned on the rendered state and text description. Intuitively, humans possess the ability to first imagine how to solve a task before executing actions. While this imagination may not always be physically accurate, it provides valuable guidance for subsequent actions.


This approach offers several advantages over previous methods. First, since the video generation model takes an image as input in addition to text, it is not constrained by the expressiveness limitations of text or code for describing complex 3D scenes. Second, as the generated video naturally represents the goal state over time in 2D space, it provides a more intuitive and informative supervision signal than alternative modalities such as embeddings~\cite{radford2021learning, sontakke2024roboclip}, text, or code. Third, the transition from the initial to the goal state is inherently continuous in the video, making the imagined goal more coherent and reasonable, akin to a chain-of-thought process. Lastly, since we use a pre-trained video generation model, no domain-specific fine-tuning is required, enabling broader generalization.

The remaining challenge is how to handle inaccuracies in the generated video.  
Some prior works~\cite{yang2024video,ko2023learning,ajay2023compositional,liang2024dreamitate, escontrela2023video, huang2024diffusion, luo2024grounding} assume that the generated videos are sufficiently accurate to reproduce action sequences. As a result, these approaches require extensive embodied training data tailored to specific tasks and scene settings to achieve satisfactory performance. 
In contrast, our method is designed for broader applicability and leverages a general-purpose video generation model~\cite{kling}. 

To address inaccuracies, we apply a two-step approach. 

First, we filter out significantly inaccurate samples, such as those that fail to follow the task description. Notably, while the video generation model we use~\cite{kling} produces generally reasonable results, it lacks consistency—a common feature of diffusion models~\cite{shen2024rethinking, huang2024diffusion}. To mitigate this, we improve the outcome by sampling multiple trials and applying a selection process to identify the most accurate sample. Constructing a selector/discriminator at key frames using a vision-language model (VLM) is significantly simpler than enhancing the capabilities of the video generation model itself, making this approach both practical and effective.  

Second, for the remaining videos—while they may not perfectly capture physical phenomena —they still provide valuable guidance, including goal images, actuator motions, grasping affordances, and more.  
Although the generated video alone is insufficient for directly reconstructing an accurate action sequence due to the inaccurate physical phenomena, it also serves as a strong guidance signal for further adjusting the policy. 
Specifically, we ensure that the generated video maintains a fixed camera position, aligning it pixel-wise with the rendered simulation states. This allows us to directly utilize 2D object segmentation masks, obtained via SAM2~\cite{ravi2024sam}, as goal supervision. Additionally, we incorporate both predicted and rendered depth as an additional supervisory signal. To further refine behavior in the simulation, we leverage GPT-4o to extract frame-wise contact information. Since multi-modal language models like GPT-4o struggle with spatial reasoning, we limit its role to determining object pairs in contact rather than estimating precise contact positions or forces.
By integrating these guidance signals, we refine the action sequence obtained from video-based pose tracking, enabling the system to effectively complete the task. Furthermore, starting from the simulation state provides a crucial advantage: we have access to the ground-truth states of the initial image, including depth and 6D pose. This information aids in pose tracking and depth estimation, further enhancing the accuracy of learned policies.  

By leveraging all these advantages, our method extends beyond simple pick-and-place tasks to tackle intricate manipulation problems where defining rewards is challenging.

We summarize our main contributions as follows:

\begin{itemize}
    \item We introduce \textit{LuciBot}, a pipeline that utilizes a general-purpose video generation model to autonomously generate supervision signals for complex embodied tasks. 
    \item We propose leveraging video generation models to provide rich semantic guidance, including goal states, motion trajectories, contact information and grasping affordances, overcoming the limitations of code-based and VLM-based rewards.
    \item We demonstrate that our approach generate guidance that leads to a reasonable policy for diverse tasks with diverse materials including rigid, articulated, elastic, plastic, granular and fluid objects.
\end{itemize}

\section{Related Works}
\label{sec:formatting}

\textbf{Generative foundation Models as Reward Functions for Robot Manipulation} Generative foundation models, trained on billions of data points, possess a strong understanding of the physical world. This makes them natural candidates for automatic and generalizable reward generation in robot manipulation tasks, eliminating the need for fine-tuning.

Previous work has explored using large language models (LLMs) as reward functions. Some approaches prompt LLMs to evaluate actions and directly parse their outputs into rewards \citep{chu2023accelerating, triantafyllidis2024intrinsic, kwon2024language}, while others use Python or domain-specific code to compute rewards \citep{ma2023eureka, xietext2reward, yu2023language, wang2023robogen}. These methods have demonstrated success in locomotion and rigid object manipulation tasks.

Similarly, large vision-language models (VLMs) have been used to generate rewards, leveraging either CLIP-style VLMs \citep{rocamonde2023vision, mahmoudieh2022zero} or more advanced models like GPT-4 \citep{wang2024rl, venuto2024code}. Some studies fine-tune VLMs on domain-specific data for reward generation \citep{guan2024task, du2023vision, yang2024trajectory, yang2024robot, sontakke2024roboclip, ma2024vision}. While fine-tuning improves performance, it often limits generalization to out-of-domain tasks.

Large video generation models, a recent development in generative foundation models, remain unexplored as reward generators for robot manipulation. Unlike text-based or code-based reward generation methods, video generation models produce frame sequences, offering continuous and semantically rich guidance. Our work investigates how these models can be applied to robot manipulation tasks.

\textbf{Video generation model for decision making}
A few studies have explored video generation models for decision-making \cite{yang2024video}. These models have been employed as reward functions for reinforcement learning \citep{escontrela2023video, huang2024diffusion, luo2024grounding}, policy models for robot manipulation \citep{liang2024dreamitate, ajay2023compositional, du2023learning, ko2023learning}, world models \citep{mendonca2023structured, zhou2024robodreamer, brooks2024video, du2023learning, yang2023learning, bruce2024genie, hong2024slowfast}, and pre-trained backbones \citep{wu2023unleashing, seo2022reinforcement, wu2023pre}.

However, most of these approaches fine-tune video generation models on domain-specific data, limiting their capability to generalize to other environments and tasks. In contrast, general-purpose video generation models, trained on vast datasets, have the potential to generalize across a wide range of tasks and materials without fine-tuning. Our work leverages a large video generation model to generate training rewards for diverse robot manipulation tasks, including soft objects, rigid objects, and articulated objects.
\section{Method}

\begin{figure*}[ht]
    \centering
    \vspace{-4mm}
    \includegraphics[width=\linewidth]{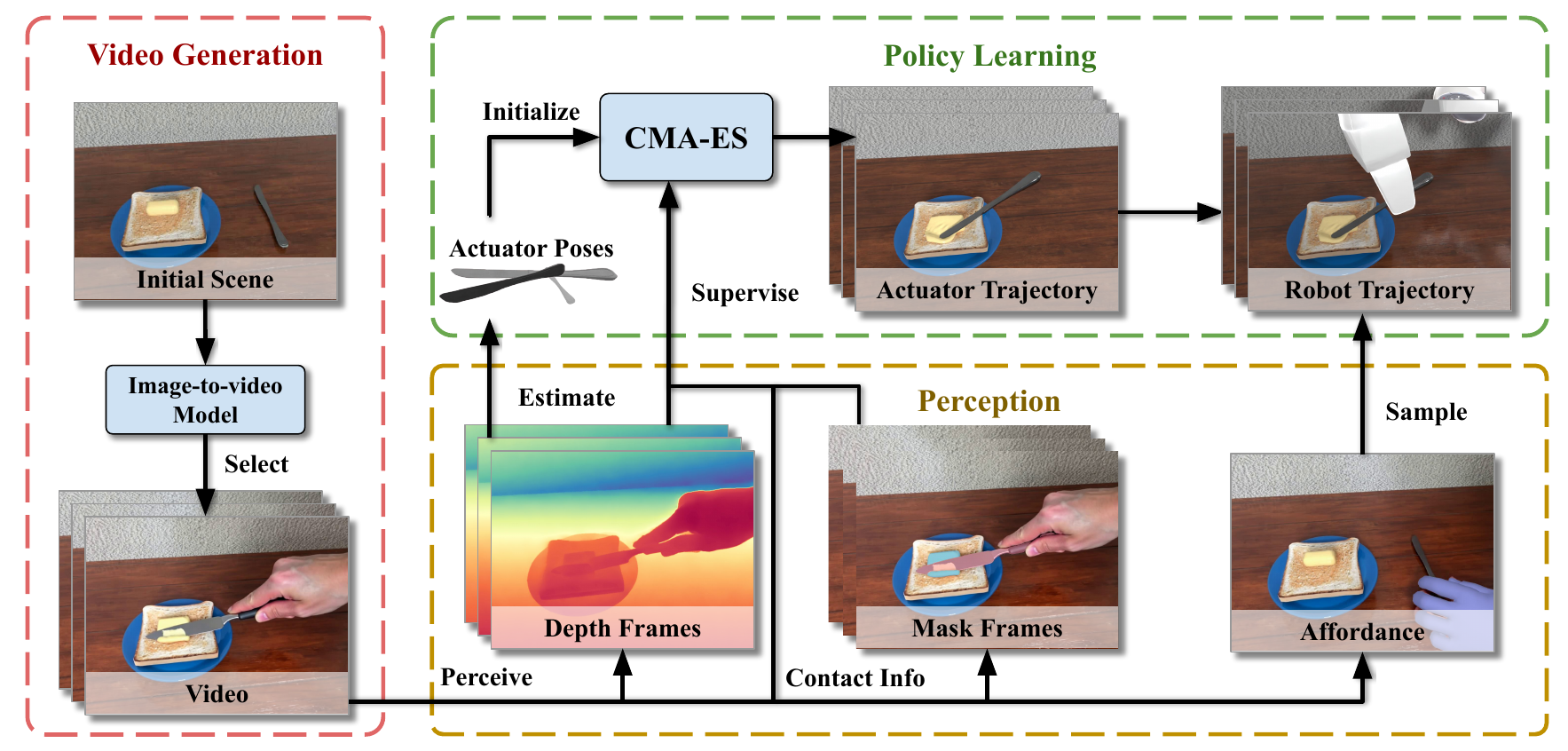}
    \caption{Our pipeline consists of three stages:
(1) Generating and selecting a valid video,
(2) Extracting various training supervision from the selected video, and
(3) Leveraging this information for policy learning.}
    \label{fig:method}
\end{figure*}

Our objective is to leverage the common sense and imagination capabilities of large video generation models to provide supervisory signals for complex embodied tasks. Given a scene configuration in the simulation and a textual task description, we generate a video based on the rendered image and text description, serving as an imagined execution process for completing the task. Supervisory signals are then extracted from this generated video to optimize an action trajectory for task execution. We will illustrate our pipeline in the following aspects: \textbf{Task Settings}, \textbf{Video Generation and Selection}, \textbf{Perception} and \textbf{Policy Learning}. 

\subsection{Task Settings}

\begin{figure*}[ht]
    \centering
    \vspace{-2mm}
    \includegraphics[width=\linewidth]{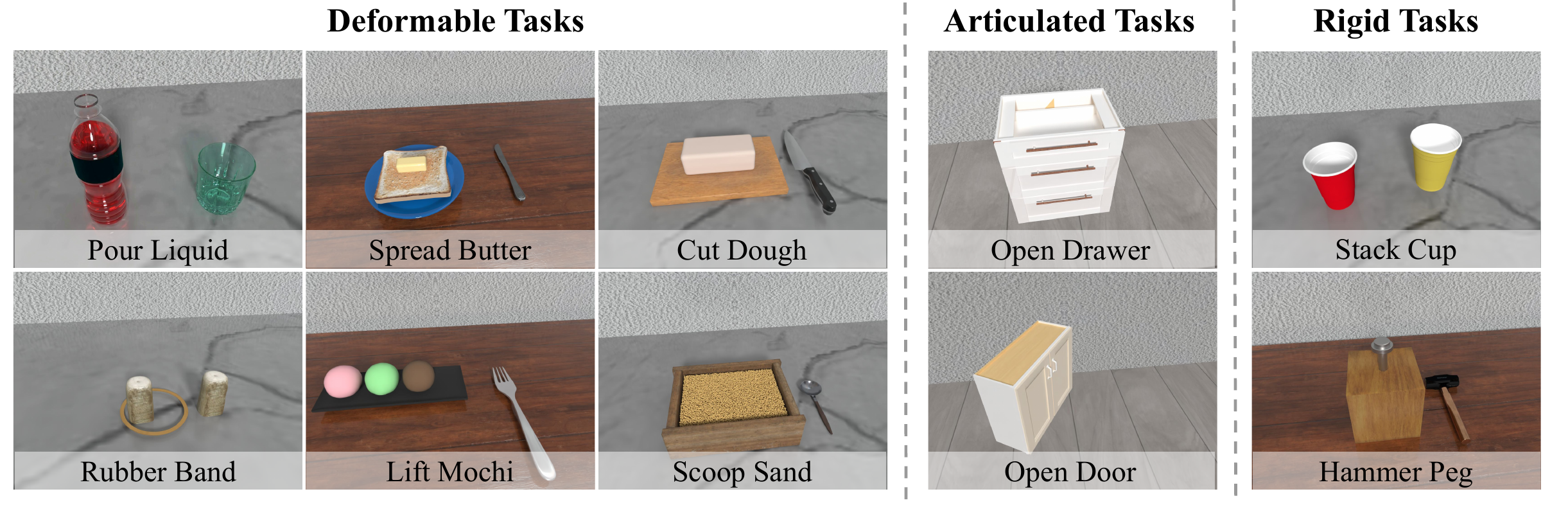}
    \caption{We propose 10 challenging tasks involving diverse materials, including \textbf{liquid, elastic, elasto-plastic, granular, articulated, and rigid objects}. Designing reward functions for those deformable tasks is inherently difficult, even with manual efforts.}
    \label{fig:task}
\end{figure*}

In general, the scene configuration is defined as $(\{O^{\text{fg}}_i\}, \{O^{\text{bg}}_i\}, T_c)$, where $\{O^{\text{fg}}_i\}$ represents the set of foreground objects (e.g., the dough and knife in Figure~\ref{fig:task}), $\{O^{\text{bg}}_i\}$ denotes the set of stationary background objects (e.g., walls and tables), and $T_c$ is the stationary camera matrix, which encodes the extrinsic and intrinsic camera parameters.

In each task, a subset of foreground objects functions as actuator objects, denoted as $O^{\text{fg}}_{\text{act}}$. These objects are actively manipulated (e.g., to be grasped by a robotic arm), such as the knife in the dough-cutting task. All other foreground objects can only be moved passively through contact with an actuator object.

Additionally, certain foreground objects serve as target objects, denoted as $O^{\text{fg}}_{\text{tar}}$, which define the primary objective of the task. Examples include the dough and butter in the tasks illustrated in Figure~\ref{fig:task}. Notably, an actuator object can also be a target object in some scenarios.


\subsection{Video Generation and Selection}

\paragraph{Video Generation} 
After setting up the scene in simulation, we render the initial scene by ray-tracing to obtain the following observations: a photorealistic image $\hat{I}_0$, ground-truth depth map $\hat{D}_0$, and segmentation mask $\hat{S}_0$. Simultaneously, we prompt a large language model (GPT-4o) to rewrite task description into a detailed text prompt for better prompting the video generation model. The image $I_0$ serves as the starting frame, and, along with the rewritten text prompt, conditions an image-to-video model~\cite{kling} to generate a video $\{I_t\}_{t=0}^T$.

\paragraph{Video Selection}
Although state-of-the-art video generation models can produce valid imaginations for soft-body and other complex manipulation tasks, they do so inconsistently. To address this, we employ a rejection sampling approach using a large pre-trained vision-language model (VLM)~\cite{achiam2023gpt} as a verifier to filter low-quality generated videos.

Specifically, we prompt the VLM to evaluates videos based on the following criteria:

\begin{itemize}   
    \item Did the generated video match the description? (2-6 points)
    \item Is the human hand motion in the video reasonable? (1-3 points)
    \item Did the scene in the video reached the goal state inferred from the description? (2-6 points)
\end{itemize}

Additionally, we prompt the VLM to detect objects not in ${O^{\text{fg}}_i}$ appearing mid-video. If any are detected, the video is assigned the lowest score.
For the detailed prompts, please refer to the supplementary materials.

\subsection{Perception}

Using previous procedure, we generate some candidate videos per task and select the highest-scoring video as the guidance video $\{I_t^G\}_{t=0}^T$. 
As demonstrated in Figure~\ref{fig:method},
we then apply a video segmentation model (SAM2~\cite{ravi2024sam}) to track foreground objects. For each foreground object $O^{\text{fg}}_i$, we use points sampled from its ground truth segmentation mask $\hat{S}_0^{O_i^{\text{fg}}}$ in the initial frame as a prompt for SAM2, which tracks $O_i^{\text{fg}}$ across the video and generates per-frame segmentation masks $\{{S_t^{O_i^{fg}}}\}_{t=0}^T$.

Next, we apply a depth inpainting model (DepthLab~\cite{liu2024depthlab}) to estimate per-frame depth maps. Assuming background objects ${O^{\text{bg}}_i}$ remain stationary, we set their depth $D_t^{O^{\text{bg}}_i}$ to their ground truth depth $\hat{D}_0^{O^{\text{bg}}_i}$. This known depth conditions DepthLab to estimate the depth of foreground objects.

To estimate foreground object poses, we use their 2D segmentation and depth information. Instead of relying on pre-trained networks such as FoundationPose, which often fail due to inaccuracies in generated videos (e.g., slight variations in object size, shape, and depth), we frame pose estimation as an optimization problem. Given a foreground object $O^{\text{fg}}_i$ at timestep $t$, we optimize for the pose that best aligns with its depth map $D_t^{O^{\text{fg}}_i}$ and 2D segmentation mask $S_t^{O^{\text{fg}}_i}$.

Finally, we sample keyframes ${I_t}, {t \in T_K}$ at a certain frequency. We prompt a VLM to identify the keyframe $I^K_{\text{start}}$ where a human hand first interacts with an object and the keyframe $I^K_{\text{end}}$ where the goal is achieved, filtering out irrelevant video segments.
Additionally, we prompt the VLM to determine whether the target object $O^{\text{fg}}_{\text{tar}}$ is in contact with any foreground objects $O^{\text{fg}}_i$ at each keyframe, providing crucial guidance for the optimization process.

\subsection{Policy Learning}


\paragraph{Supervisions}

As shown in Figure~\ref{fig:method}, we extract supervisory signals from the generated video to guide training:


\textbf{2D Mask of the Actuator.}
We compute the Intersection over Union (IoU) between the actuator's 2D segmentation mask in simulation ($\hat{S}t^{O_{\text{act}}^{\text{fg}}}$) and its segmentation mask extracted from the video ($S_t^{O_{\text{act}}^{\text{fg}}}$) for all keyframe timesteps $t \in T_K$. This prevents the motion trajectory drifting too far from the video.

\textbf{2D Mask of the Target Object.}
Similarly, we compute the IoU between the target object's 2D segmentation mask in simulation ($\hat{S}_{\text{end}}^{O^{\text{fg}}_{\text{tar}}}$) and its segmentation mask extracted from the video ($S_{\text{end}}^{O^{\text{fg}}_{\text{tar}}}$) as a reward signal to guide task completion. This applies only to the final keyframe.

\textbf{3D Point Cloud of the Actuator.}
Using rendered simulator depth ($\hat{D}_{\text{end}}^{O^{\text{fg}}_{\text{act}}}$) and predicted depth from the video ($D_{\text{end}}^{O^{\text{fg}}_{\text{act}}}$) in the final keyframe, we project the actuator's depth into 3D point clouds and compute the Chamfer distance between simulation and video-extracted point clouds.

\textbf{3D Point Cloud of the Target Object.}
Analogous to the actuator, we compute the Chamfer distance for the target object’s 3D point clouds as another supervisory signal.

\textbf{Contact Information.}
Contact is crucial in manipulation tasks. To enforce physical interaction, we introduce a penalty when an object and target object are not in contact (if they should be in contact). For example, in Figure~\ref{fig:method}, the knife contacts the butter from the second frame onward, ensuring realistic interaction constraints.

\textbf{Affordance.}
The grasping position of human hand in the generated video provides guidance for the robot arm gripper about where to grasp an object. We detect human hand pose in the first keyframe $I^K_{\text{start}}$ using an off-the-shelf hand detector (WiLoR~\cite{potamias2024wilor}), and label the pixels around finger tips as the affordance area. 

\paragraph{Trajectory Optimization and Robot Manipulation}

To simplify the problem, we optimize the motion trajectory of the actuator directly in the simulator, abstracting the action sequence into a series of waypoints based on the frequency of selected keyframes. We initialize the action sequence using the actuator's estimated poses from these keyframes and employ a sample-based optimization method, CMA-ES, to refine the trajectory for task completion.

Finally, we sample grasping poses based on grasping affordance and employ inverse kinimatics and path-planning algorithms to control the robot arm to grasp the actuator and follow this trajectory.

\section{Experiment}

\begin{table*}[ht]
    \centering
    \begin{tabular}{l|cccccccc|c}
        \toprule
         & \textbf{liquid} & \textbf{dough} & \textbf{butter} & \textbf{sand} & \textbf{rubber} & \textbf{mochi} & \textbf{peg} & \textbf{cup} & \textbf{Avg}\\
        \midrule
        \textbf{VLM-RMs}~\cite{rocamonde2023vision}    & 4.84 & 2.84 & 4.68 & 5.11 & 5.00 & 4.36 & 3.42 & 2.79 & 4.13\\
        \textbf{T2R}~\cite{xietext2reward, yu2023language}        & 4.11 & 4.79 & 3.16 & 2.89 & 3.63 & 3.26 & 4.84 & 3.37 & 3.76\\
        \textbf{AVDC}~\cite{ko2023learning}              & 5.79 & 3.32 & 6.00 & 3.53 & / & 6.00 & 5.42 & 6.00 & -\\
        \midrule
        \textbf{Ours w/o Init}      & 2.74 & 3.63 & 2.16 & 3.53 & 2.79 & 3.16 & 2.47 & 3.37 & 2.98\\
        \textbf{Ours w/o Contact}   & 2.05 & 5.42 & 3.52 & 4.95 & 2.58 & 3.21 & 3.11 & 4.11 & 3.62\\
        \midrule
        \textbf{Ours}              & \textbf{1.47} & \textbf{1.00} & \textbf{1.47} & \textbf{1.00} & \textbf{1.00} & \textbf{1.00} & \textbf{1.74} & \textbf{1.37} & \textbf{1.26}\\
        \bottomrule
    \end{tabular}
    \caption{Scores for GPT-ranking. Lower score indicating better performance.}
    \label{tab:rank}
\end{table*}

\begin{table*}[ht]
    \centering
    \begin{tabular}{l|cccccccc|c}
        \toprule
         & \textbf{liquid} & \textbf{dough} & \textbf{butter} & \textbf{sand} & \textbf{rubber} & \textbf{mochi} & \textbf{peg} & \textbf{cup} & \textbf{Avg}\\
        \midrule
        \textbf{VLM-RMs}~\cite{rocamonde2023vision}    & 0.00 & 0.79 & 0.01 & 0.00 & 0.01 & 0.34 & 0.95 & 0.68 & 0.35\\
        \textbf{T2R}~\cite{xietext2reward, yu2023language}      & 0.00 & 0.07 & 0.56 & 0.68 & 0.46 & 0.49 & 0.98 & 0.59 & 0.48\\
        \textbf{AVDC}~\cite{ko2023learning}              & 0.01 & 0.88 & 0.00 & 0.99 & $/$ & 0.00 & \textbf{1.00} & 0.00 & -\\
        \midrule
        \textbf{Ours w/o Init}      & 0.52 & 0.16 & 0.71 & 0.15 & 0.95 & 0.65 & 0.95 & 0.51 & 0.58\\
        \textbf{Ours w/o Contact}   & 0.96 & 0.10 & 0.01 & 0.00 & 0.91 & 0.93 & 0.99 & 0.45 & 0.51\\
        \midrule
        \textbf{Ours}               & \textbf{1.00} & \textbf{1.00} & \textbf{0.99} & \textbf{1.00} & \textbf{1.00} & \textbf{1.00} & 0.88 & \textbf{0.96} & 0.98\\
        \bottomrule
    \end{tabular}
    \caption{Scores for VQA-score using GPT-4o model. Higher score indicating better performance.}
    \label{tab:vqascore}
\end{table*}

\paragraph{Task Settings}
We evaluate our method on 10 benchmark tasks spanning diverse materials (as shown in Figure~\ref{fig:task}), including fluid, elasto-plastic, granular, elastic, and rigid objects, thanks to the powerful simulator Genesis \cite{Genesis}. These tasks are categorized into deformable tasks, articulated tasks, and rigid tasks.

\noindent\textbf{Deformable Tasks}
\begin{itemize}
    \item \textbf{Cut Dough}: Use a knife to cut a dough.  
    \textit{(Actuator: knife, Target: dough)}
    \item \textbf{Pour Liquid}: Pour liquid from a plastic bottle into a glass.  
    \textit{(Actuator: plastic bottle, Target: liquid)}
    \item \textbf{Rubber Band}: Stretch a rubber band onto two wooden sticks. No rigid actuator exists; a set of particles on the band is used as the actuator.  
    \textit{(Actuator: selected particles, Target: rubber band)}
    \item \textbf{Spread Butter}: Use a butter knife to spread butter on bread.  
    \textit{(Actuator: butter knife, Target: butter)}
    \item \textbf{Lift Mochi}: Lift a brown mochi using a fork.  
    \textit{(Actuator: fork, Target: brown mochi)}
    \item \textbf{Scoop Sand}: Scoop a spoonful of sand with a spoon.  
    \textit{(Actuator: spoon, Target: sand)}
\end{itemize}

\noindent\textbf{Rigid Tasks}
\begin{itemize}
    \item \textbf{Hammer Peg}: Drive a peg into a wooden block using a hammer.  
    \textit{(Actuator: hammer, Target: peg)}
    \item \textbf{Stack Cups}: Stack a yellow cup on a red cup.  
    \textit{(Actuator: both cups, Target: both cups)}
\end{itemize}

\noindent\textbf{Articulated Tasks}
\begin{itemize}
    \item \textbf{Open Door}: Rotate a door to open it.  
    \textit{(Actuator: door, Target: door)}
    \item \textbf{Open Drawer}: Pull a drawer to open it.  
    \textit{(Actuator: drawer, Target: drawer)}
\end{itemize}

\paragraph{Implementation}
We leverage Genesis~\cite{Genesis} to simulate diverse materials, manually tuning material parameters for accuracy. For fluid simulation in the \textbf{Pour Liquid} task, we use an SPH solver, while an MPM solver is applied to all other deformable tasks. We define material properties as follows: \textit{ElastoPlastic} for butter, dough, and mochi; \textit{Elastic} for the rubber band; and \textit{Sand} for granular materials. To enhance simulation stability, we set the mass of all rigid objects to 0.1. In video processing, we use SAM2~\cite{ravi2024sam} for video segmentation, DepthLab~\cite{du2020depthlab} for depth estimation and WiLoR~\cite{potamias2024wilor} for hand detection.
For tasks with rigid actuators, we optimize a sequence of 6-DoF waypoints and use an embedded PD controller to transition between waypoints. In the \textbf{Rubber Band} task, where no rigid actuator exists, we directly set the velocity of selected particles to guide them toward the next waypoint.  
We employ an open-source CMA-ES implementation~\cite{hansen2019pycma}, setting the population size to 128. The best result within 5 iterations (a total of 640 samples) is selected as the final optimized trajectory.  
We run multiple trials for all baselines, and display the best of each baseline in the qualitative results.

\vspace{-5mm}

\paragraph{Evaluation Metrics}  
In most tasks, defining a precise success indicator or value function to measure task completion is challenging, as such criteria are difficult to construct—a limitation our method aims to overcome. Consequently, we adopt evaluation metrics commonly used in generative methods~\cite{jiang2023llm,lin2024evaluating,buzhinsky2023metrics}, specifically \textbf{GPT-ranking}~\cite{jiang2023llm} and \textbf{VQA-score}~\cite{lin2024evaluating}.  
For \textbf{GPT-ranking}, we collect the resulting images from our method and baseline approaches, then rank them using GPT-4o, reporting the average ranking. For \textbf{VQA-score}, we evaluate the text-image alignment by measuring how well the final state descriptions correspond to the generated results. In our case, we use the GPT-4o model for more accurate evaluation.

\begin{figure*}[ht]
    \centering
    \includegraphics[width=\linewidth]{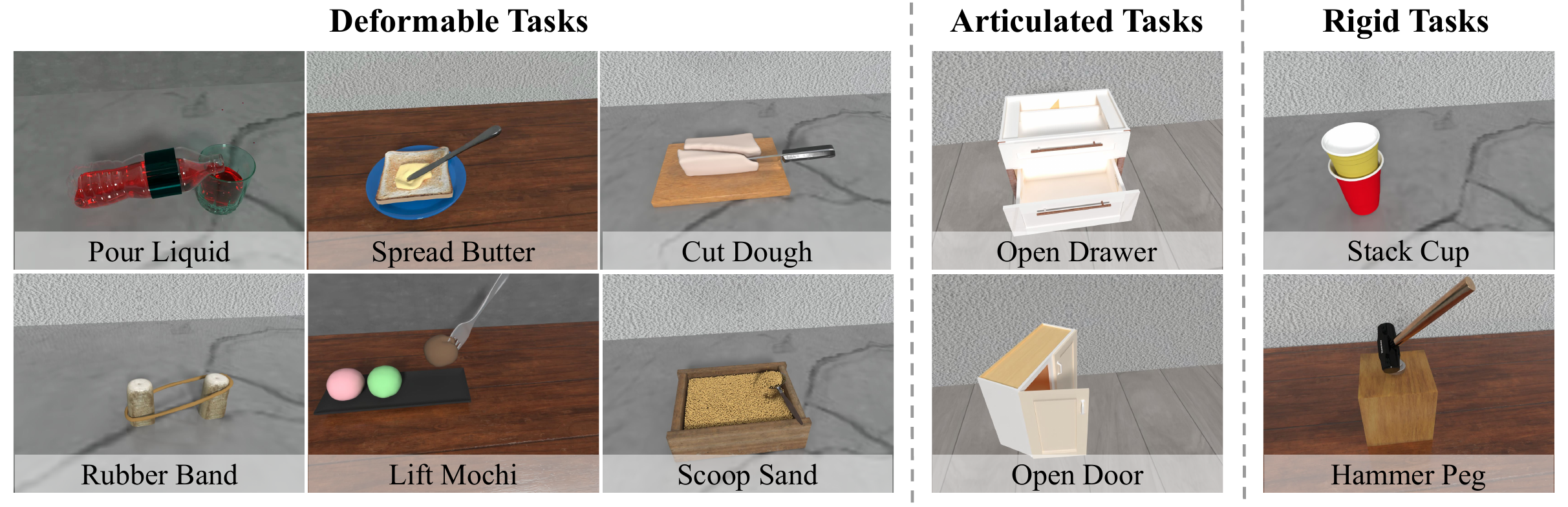}
    \vspace{-2mm}
    \caption{We display the end frames executing our policy, demonstrating that our method can reasonably perform these tasks.}
    \vspace{-2mm}
    \label{fig:visualization}
\end{figure*}

\begin{figure*}[h]
    \centering
    \includegraphics[width=\linewidth]{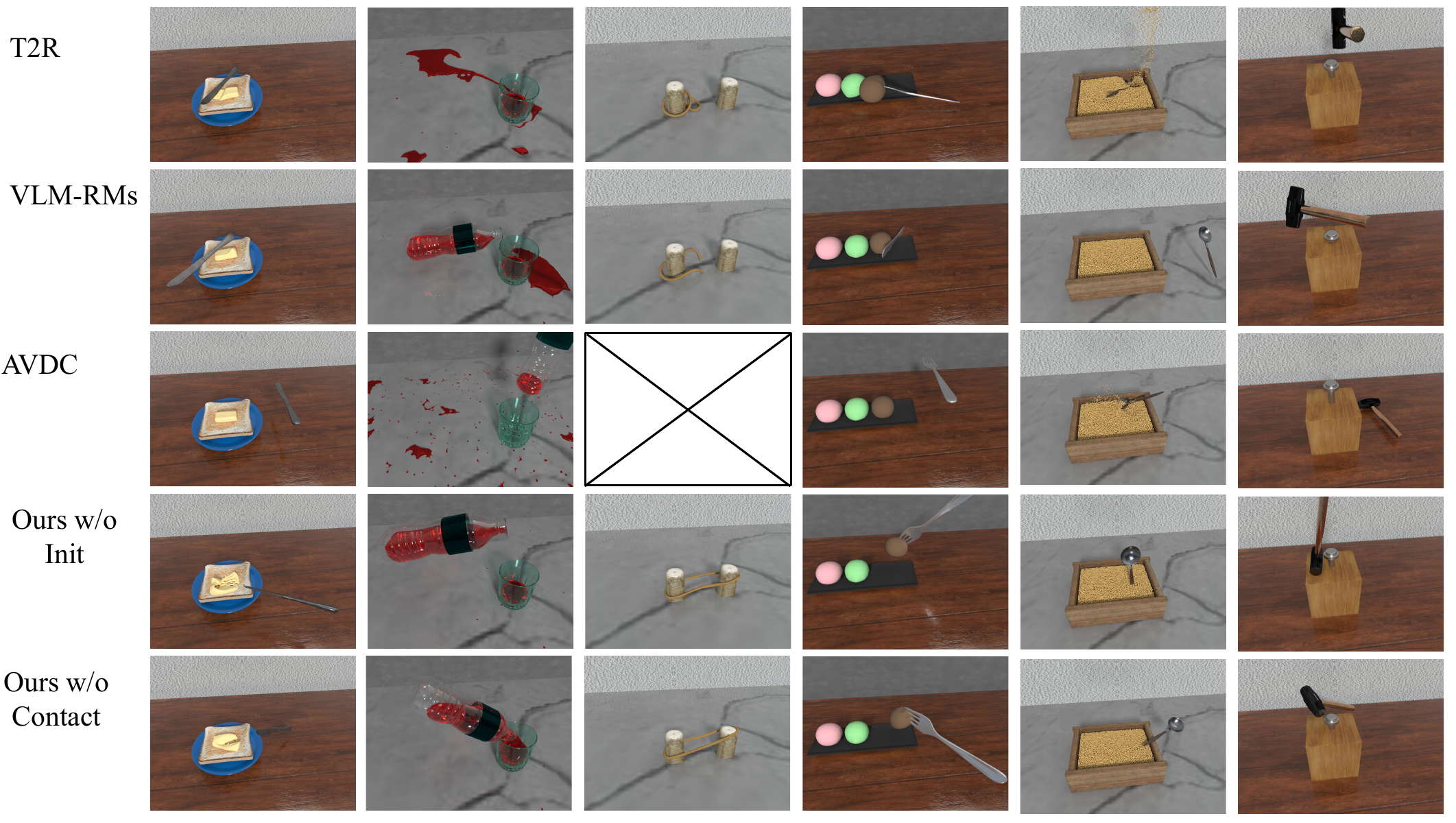}
    \caption{We show the best qualitative results for all baselines of multiple trials.}
    \label{fig:more1}
\end{figure*}

\subsection{Baselines}  

\subsubsection{Implementation}
We compare our approach with various reward generation and video-to-policy methods, including prior works on vision-language model-based reward models (VLM-RMs)~\cite{rocamonde2023vision, mahmoudieh2022zero}, text-to-reward (T2R)~\cite{xietext2reward, yu2023language}, and AVDC~\cite{ko2023learning}.  

For \textbf{VLM-RMs}~\cite{rocamonde2023vision}, representing reward generation via vision-language models, we follow its implementation by rendering the final frame and computing CLIP features to derive rewards.  

For \textbf{T2R}~\cite{xietext2reward, yu2023language}, we prompt GPT-4o with relevant APIs from the Genesis simulator to generate a reward function for the final state. Detailed prompts are provided in Section~\ref{sec:prompt}.  

For \textbf{AVDC}~\cite{ko2023learning}, we adjust it to fit in our setting by estimating object poses using optical flow with the generated video as input, and directly follow the estimated pose sequence to execute the task. Note that we are not comparing to the full version of AVDC where the video generation model is specially tuned. It can be viewed as an ablated version without further optimization process. We don't apply this baseline on the \textbf{Rubber Band} task since there's no rigid actuator to track.

Additionally, inspired by RL-VLM-F \cite{wang2024rl}, we implement a preference-based scoring method (\textbf{VLM-Pref}). Specifically, we perform an initial ranking of the first batch of samples using GPT-4o via a quicksort-based approach, assigning scores uniformly from 0 to 50. For each new sample, we conduct a binary search among previously ranked samples and assign a score based on its relative position. Due to the high computational cost and high financial cost of this ranking method (requiring too many API calls), we only evaluate it on two tasks, as detailed in Table~\ref{tab:pref}.  

We evaluate baselines on 8 tasks, excluding articulated object cases. In articulated tasks, actions are directly applied to the actuator, which is itself the articulated object, making execution straightforward by simply setting the degrees of freedom (DoFs) to the estimated positions.  

\subsubsection{Analysis}
As shown in the quantitative results in Table~\ref{tab:rank} and Table~\ref{tab:vqascore}, \textbf{AVDC} ranks the lowest in most cases. This highlights the difficulty of achieving precise pose tracking, given that both the generated video and the depth map predicted from it lack sufficient accuracy.  

\begin{figure*}[htbp]
    \centering
    \vspace{-2mm}
    \includegraphics[width=\linewidth]{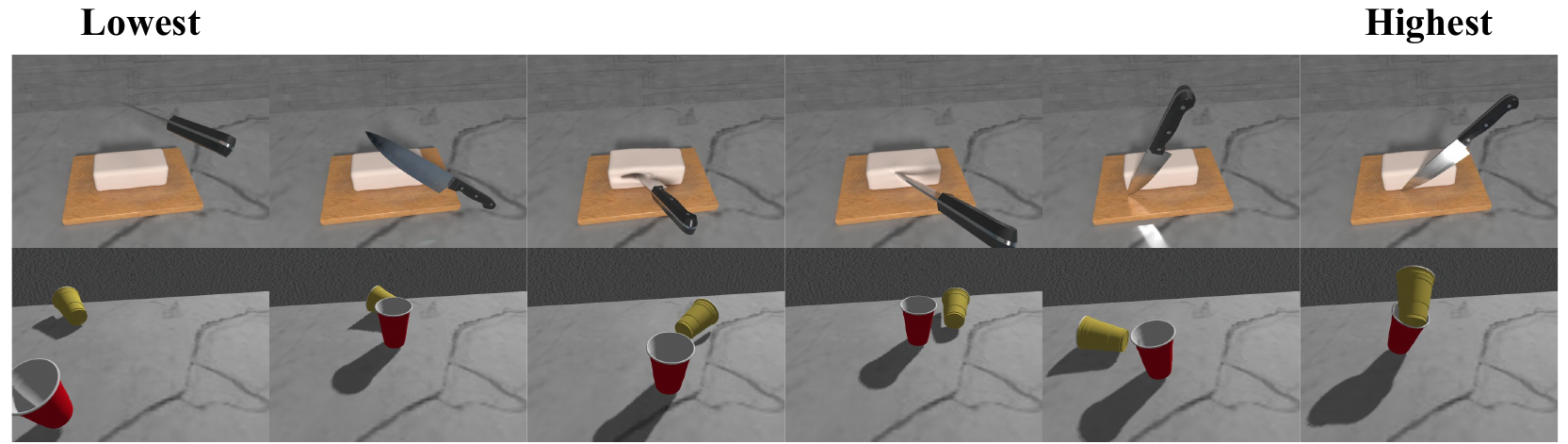}
    \caption{We present the ranking results of \textbf{VLM-Pref}, with images uniformly sampled from the ranking.}
    \label{fig:rlvlmf}
\end{figure*}

The performance of \textbf{T2R} and \textbf{VLM-RMs} is reasonable in simpler tasks but fails in most cases. T2R performs well in tasks such as \textbf{Lift Mochi} and \textbf{Hammer Peg}, where goals can be accurately defined using code APIs. However, it struggles in tasks where the goal state is harder to describe programmatically or is unknown yet. For instance, in \textbf{Pour Liquid}, the reward function encourages maximizing the amount of liquid within the bounding box of the glass but fails to penalize spillage. Similarly, in \textbf{Scoop Sand}, the reward encourages maximizing the amount of sand within the bounding box of the spoon, leading to unintended behaviors such as the policy increasing the spoon’s bounding box by tilting it at an unnatural angle within the sandbox.  
\textbf{VLM-RMs} provide some degree of effectiveness across all tasks, but their scoring mechanism is often too ambiguous. The scores fail to capture fine-grained differences when the rendered states appear similar. For example, in the \textbf{Scoop Sand} task, it scores the highest when the spoon is far from the sandbox. For \textbf{VLM-Pref}, we observe that while GPT-4o can judge task completion, it struggles as a continuous value function for measuring temporal progress. When ranking hundreds of data samples in pairs, GPT-4o fails to maintain long-term consistency, as shown in Figure~\ref{fig:rlvlmf}.

Our method significantly outperforms all baselines across tasks, as the goal imagined by the video generation model provides a more direct and accurate supervision signal. For example, in \textbf{Pour Liquid}, our method effectively filters out cases where water spills outside the glass, as there is no 2D mask corresponding to liquid outside the container, ensuring a more robust training signal.  

\begin{table}[ht]
    \centering
    \begin{tabular}{l|cc}
        \toprule
         & \textbf{Stack Cup} & \textbf{Cut Dough} \\
        \midrule
        \textbf{VLM-Pref} & 2.0 / 0.18 & 2.0 / 0.62 \\
        \textbf{Ours} & \textbf{1.0 / 0.96} & \textbf{1.0 / 1.00}\\
        \bottomrule
    \end{tabular}
    \vspace{-2mm}
    \caption{Quantitative comparison with \textbf{VLM-Pref} method. We show both scores in  GPT-ranking / VQA-score format.}
    \vspace{-5mm}
    \label{tab:pref}
\end{table}

\subsection{Ablations}

\begin{figure*}[htbp]
    \centering
    \vspace{-2mm}
    \includegraphics[width=\linewidth]{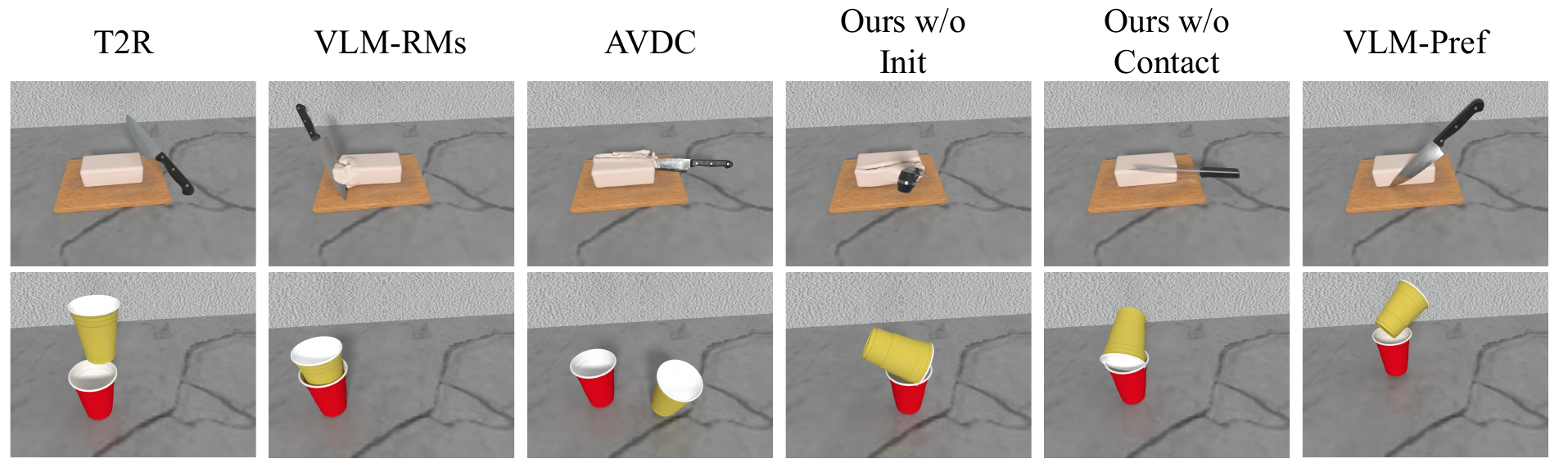}
    \caption{We show the best qualitative results for all baselines of multiple trials. These two tasks include \textbf{VLM-Pref} baseline.}
    \label{fig:more2}
\end{figure*}

We conduct ablation studies on two crucial components of our pipeline: \textbf{Contact Information} and \textbf{Trajectory Initialization}.  

To analyze their impact, we compare the following ablated versions:  

\begin{itemize}  
    \item \textbf{Ours w/o Contact}: This variant excludes contact information from the optimization process, preventing it from guiding trajectory search.  
    \item \textbf{Ours w/o Init}: This variant omits trajectory initialization using the tracked 6D pose, instead relying on random initialization for waypoints.  
\end{itemize}  

For \textbf{Ours w/o Contact}, we observe a more significant performance drop compared to \textbf{Ours w/o Init}, indicating that contact information plays a crucial role in these tasks. Contact constraints effectively reduce the search space to a valid subset with correct object interactions, improving sample efficiency and minimizing cases where the result fits the 2D masks but fails due to incorrect collisions.  

For \textbf{Ours w/o Init}, we observe a performance decline; however, it still generally outperforms the other baselines. This demonstrates that while initialization is beneficial, it is not strictly necessary for success. Although the estimated trajectory is not entirely accurate (as seen in the AVDC baseline), it provides a useful directional prior or motion pattern that aids in task completion. 

We present the best final outcomes from multiple trials for each baseline and ablated version, displaying their corresponding end frames in Figure~\ref{fig:more1} and Figure~\ref{fig:more2}.  

From these results, we observe that baseline methods struggle to solve most of the listed tasks. \textbf{T2R} succeeds only in the \textbf{Hammer Peg} task, and \textbf{VLM-RMs} successfully complete only the \textbf{Hammer Peg} and \textbf{Stack Cup} tasks. The direct tracking approach employed by \textbf{AVDC} fails to achieve success in any task, and \textbf{VLM-Pref} also fails in the two tasks we evaluated, highlighting the complexity of these tasks and the necessity of careful reward design.  

The ablated versions perform well across multiple tasks, including \textbf{Spread Butter}, \textbf{Pour Liquid}, \textbf{Rubber Band}, \textbf{Lift Mochi}, and \textbf{Hammer Peg}, further demonstrating the effectiveness of our approach.  

\subsection{Real-World Evaluations}

Beyond rendered images, our method can also be applied to real-world scenarios. Specifically, we first set up a real-world scene, capture an image of it, and generate a video conditioned on the image and task description.  

Next, we build a digital twin of the real-world environment within the simulator, ensuring alignment with the same camera parameters. We then execute the remainder of our pipeline in simulation and transfer the optimized trajectory back to the real-world for task execution. 

Resetting the scene for each sample, especially in scenarios involving liquids, is highly labor-intensive, making simulation-based optimization a more practical approach than sampling in real-world.  

\begin{figure}[ht]
    \centering
    \vspace{-2mm}
    \includegraphics[width=\linewidth]{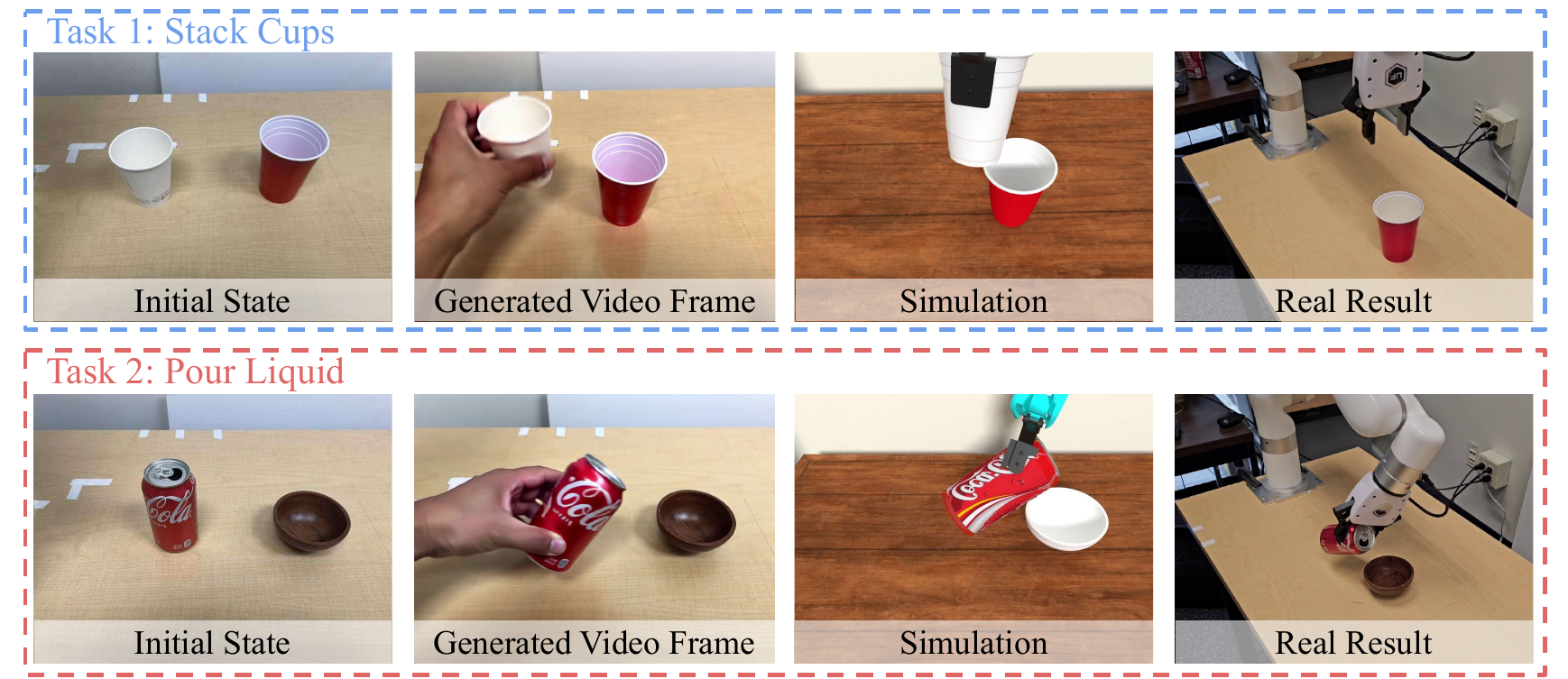}
    \vspace{-4mm}
    \caption{We present two real-world cases similar to the Stack Cups and Pour Liquid tasks. The images from left to right show the initial real-world state, a frame from the generated video, a frame from the simulated digital twin, and the final real-world outcome.}
    \vspace{-4mm}
    \label{fig:real_world}
\end{figure}

The results in Figure~\ref{fig:real_world} show that our method can be successfully transferred to the real-world.

\subsection{Video Selection}

\begin{figure}[ht]
    \centering
    \vspace{-2mm}
    \includegraphics[width=\linewidth]{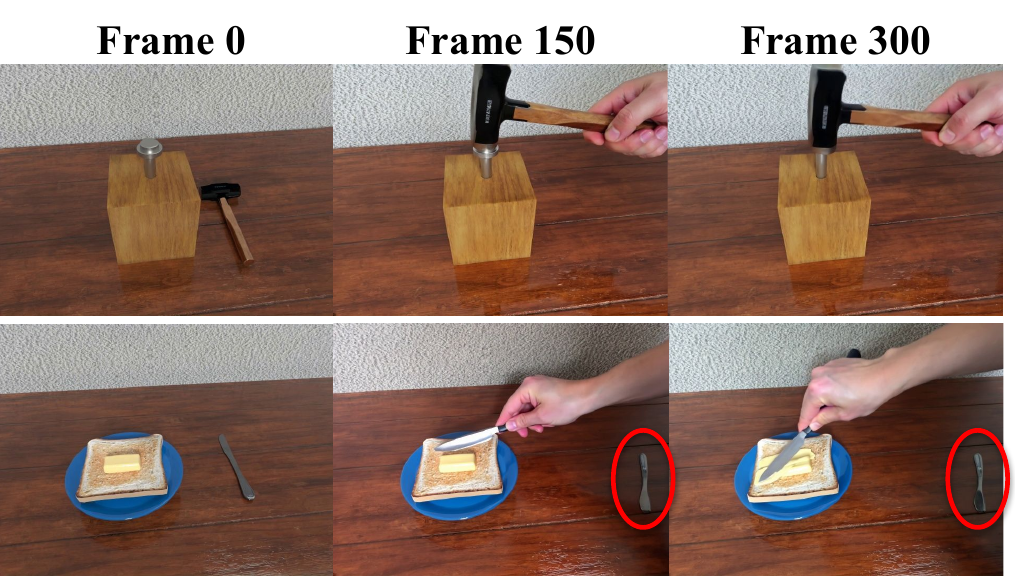}
    \vspace{-4mm}
    \caption{Examples of videos rejected by the verifier}
    \vspace{-2mm}
    \label{fig:verifyer}
\end{figure}

\begin{figure}[htbp]
    \centering
    \includegraphics[width=\linewidth]{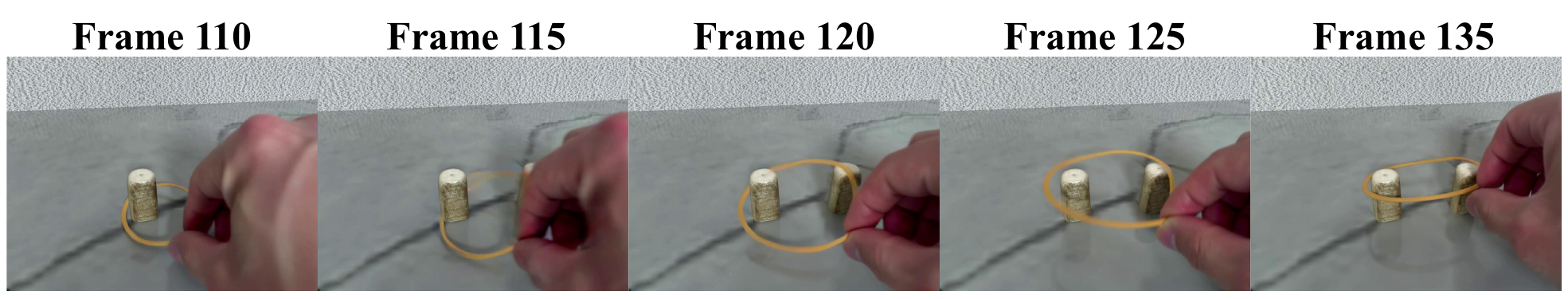}
    \vspace{-2mm}
    \caption{Some frames of the generated guidance video for task \textbf{rubber band}.}
    \vspace{-4mm}
    \label{fig:rubber}
\end{figure}

The video selection process automates the pipeline, eliminating the need for manual video selection. To evaluate its effectiveness, we conduct an experiment using a set of 62 generated videos, where some are usable (support successful policy learning) while others do not. Each videos are labeled accordingly.

We apply our verifier prompt to all videos, classifying those with scores above 12 as usable. Precision, recall, and F1 score are used as evaluation metrics, achieving 0.8, 0.8333, and 0.8163, respectively.

Figure~\ref{fig:verifyer} presents examples of videos rejected by the verifier. In the first case, although a hand picks up the hammer and strikes the peg, it fails to drive the peg fully into the wooden block. Consequently, it receives a low score for the third evaluation criterion: Did the scene in the video reach the goal state inferred from the description? In the second case, a new knife is used to spread butter instead of the one on the table, violating the rule that no new objects should appear mid-video, leading to its immediate rejection.

While a physically accurate video simplifies trajectory optimization, we also demonstrate that even videos with physical inaccuracies can still facilitate successful trajectory optimization. For example, Figure~\ref{fig:rubber} presents frames from a video used to guide the rubber band task. While the physics of the hand pulling the rubber band is not correctly depicted, the final goal state remains accurate. In this case, the goal state alone provides sufficient supervision for successful trajectory optimization.

Experimental results demonstrate that, despite the inherent instability of video generation, our verifier effectively filters out low-quality videos, enhancing the robustness of our proposed method.

\section{Conclusion}

We proposed LuciBot, a pipeline that extracts training supervision for embodied tasks from videos generated by large pre-trained video generation models. Leveraging the imagination capability of recent video generation models, LuciBot first renders an image from the simulation environment and queries the model to generate a video as an imaginary demonstration. Rich and diverse information is then extracted from the generated video, including 6D object poses, 2D segmentation masks, depth maps, contact information, and affordance information. Finally, the extracted supervision is used for trajectory optimization to complete the embodied task.

Experimental results demonstrate that LuciBot effectively generates training supervision for challenging embodied tasks, whereas previous reward generation methods fail due to their limitations. Additional experiments validate the contributions of individual components in LuciBot and its potential for real-world applications.

Since state-of-the-art pre-trained video generation models are limited in video length, our work primarily focuses on short-horizon tasks. A promising future direction is extending training supervision to long-horizon embodied tasks, such as cleaning an entire dining table. However, this presents a challenge, as longer videos accumulate errors over time.

\nocite{chen2021learning}
\nocite{yang2024spatiotemporal}

{
    \small
    \bibliographystyle{ieeenat_fullname}
    \bibliography{main}
}

\clearpage
\section{Appendix}
\maketitle
\appendix

\section{Prompting details}
\label{sec:prompt}

\noindent
\begin{minipage}{\textwidth}
\noindent
Prompts for video evaluation:
\begin{lstlisting}[style=markdownstyle, basicstyle=\normalsize\ttfamily]
You are an assistant that helps me evaluate AI generated videos. I will give you a few key frames of a video, and a description used to generate the given video. You will rate the given video according to the following criterion:
    (1) Did the generated video match the description?
        6 points: The video strictly follows each step in the description.
        4 points: The video follows parts of the description, but not all.
        2 points: The video fails to follow the description

    (2) Is the human hand motion in the video reasonbale?
        3 points: The video includes a human hand with a natural structure. Its interaction with other objects (such as grip position and gestures) follows physics and common sense without distortion.
        2 points: The video includes a human hand, but either its structure appears unnatural or its interaction with objects is unrealistic.
        1 point:  There is no human hand in the video.

    (3) Did the scene in the video reached the goal state inferred from the description? 
        6 points: The scene ultimately reaches the goal state.
        4 points: The scene reaches the goal state at some point but later moves away from it (for example if a task demands a door to be open, and the door is opened and then is closed again).
        2 points: The scene never reaches the goal state.

After giving a detailed analysis, please summarize your answer in the following format:
```video_evaluation
    score: the score for the video (in the format of x/15)
```

Remember:
(1) Do not hallucinate, your answer should only base on the provided frames
(2) Do not simply copy the task description, you should analyze what is happening in the frames 
\end{lstlisting}
\end{minipage}
\hfill
\twocolumn

\noindent
\begin{minipage}{\textwidth}
\noindent
Prompts for text to reward baseline:
\begin{lstlisting}[style=markdownstyle, basicstyle=\normalsize\ttfamily]
You are a reward engineer trying to write reward functions to solve reinforcement learning tasks as effective as possible. 
Your goal is to write a reward function for the environment that will help the agent learn the task described in text. 
Your reward function should use useful variables from the environment as inputs. As an example, the reward function signature can be:
def compute_reward() -> float:
    ...
    return reward

The code output should be formatted as a python code string: "
python ...
".

Some helpful tips for writing the reward function code: 
    (1) You may find it helpful to normalize the reward to a fixed range by applying transformations like numpy.exp to the overall reward or its components
    (2) If you choose to transform a reward component, then you must also introduce a temperature parameter inside the transformation function; this parameter must be a named variable in the reward function and it must not be an input variable. Each transformed reward component should have its own temperature variable
    (3) Make sure the type of each input variable is correctly specified; a float input variable should not be specified as numpy.array
    (4) Most importantly, the reward code's input variables must contain only attributes of the provided environment definition. Under no circumstance can you introduce new input variables.
    (5) The task contains deformable objects and rigid objects as actuator or static backgrounds. 
    (6) For a rigid object named obj, there are the following methods:
        obj.get_qpos().detach().cpu().numpy() -> return the 7d pose (pos and quat) for it
        obj.get_vel().detach().cpu().numpy() -> return the linear velocity
        obj.get_ang().detach().cpu().numpy() -> return the angular velocity
        obj.get_AABB().detach().cpu().numpy() -> return the axis aligned bounding box
    (7) For a deformable object named obj, there are the following methods:
        obj.get_particles() -> return the positions of particles (the deformable objects are always represented by particles)
\end{lstlisting}

\end{minipage}

\hfill
\twocolumn

\begin{figure*}[ht]
    \centering
    \vspace{-2mm}
    \includegraphics[width=\linewidth]{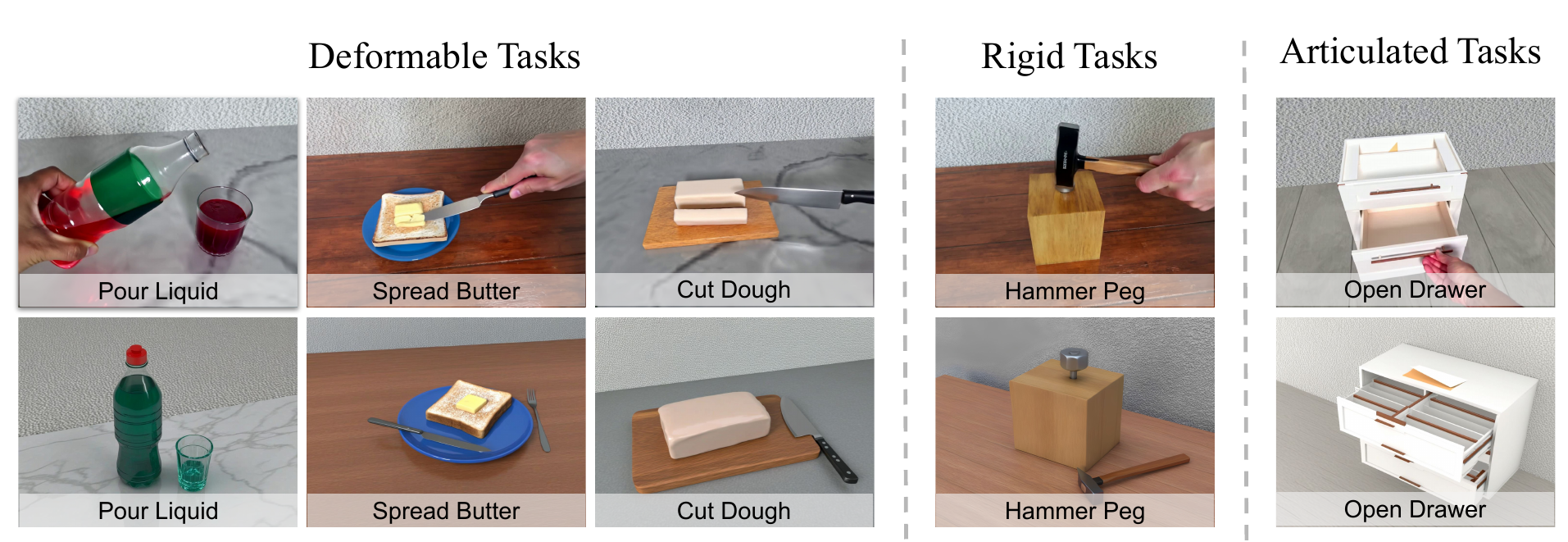}
    \caption{We present the comparison between goal images obtain via video generation model and image editing model. Results of video generation are presented in the first row. Results of image generation are presented in the second row. }
    \label{fig:img_goal}
\end{figure*}

\section{Goal image generation with image editing}

An alternative approach to generating goal images—a crucial, if not the most important, component of training supervision—is leveraging an image editing model\cite{kling_img}. In Fig.~\ref{fig:img_goal}, the goal images in the second row are generated using the initial simulation frame as a reference along with a text prompt. The results clearly show that this approach fails to preserve the correct scene layout, making it unsuitable for generating accurate goal images.

\end{document}